\documentclass[runningheads]{llncs}
\usepackage{graphicx}
\usepackage{amsmath,amssymb} % define this before the line numbering.
\usepackage{color}
\usepackage{tabularx}
\usepackage{multirow}
\usepackage{diagbox}
\usepackage{hyperref}
\usepackage{indentfirst}

\usepackage{algorithm}
\usepackage{algpseudocode}
\makeatletter
\newcommand{\algmargin}{\the\ALG@thistlm}
\makeatother
\newlength{\whilewidth}
\algdef{SE}[parWHILE]{parWhile}{EndparWhile}[1]
  {\parbox[t]{\dimexpr\linewidth-\algmargin}{%
     \hangindent\whilewidth\strut\algorithmicwhile\ #1\ \algorithmicdo\strut}}{\algorithmicend\ \algorithmicwhile}%
\algnewcommand{\parState}[1]{\State%
  \parbox[t]{\dimexpr\linewidth-\algmargin}{\strut #1\strut}}

\makeatletter
\def\thickhline{%
  \noalign{\ifnum0=`}\fi\hrule \@height \thickarrayrulewidth \futurelet
   \reserved@a\@xthickhline}
\def\@xthickhline{\ifx\reserved@a\thickhline
               \vskip\doublerulesep
               \vskip-\thickarrayrulewidth
             \fi
      \ifnum0=`{\fi}}
\makeatother
\newlength{\thickarrayrulewidth}
\newcolumntype{Y}{>{\centering\arraybackslash}X}
\begin{document}
\pagestyle{headings}
\mainmatter

\def\ACCV20SubNumber{967}  % Insert your submission number here

%===========================================================
\title{MatchGAN: A Self-Supervised Semi-Supervised Conditional Generative Adversarial Network} % Replace with your title
\titlerunning{MatchGAN}
\author{Jiaze Sun\inst{1}\and %\orcidID{0000-0002-5846-0056} \and
Binod Bhattarai\inst{1}\and %\orcidID{???} \and
Tae-Kyun Kim\inst{1,2}} % \orcidID{???}}
\authorrunning{J. Sun et al.}
\institute{Imperial College London, Exhibition Road, London SW7 2AZ, UK \and
Korea Advanced Institute of Science and Technology, 291 Daehak-ro, Yuseong-gu, Daejeon 34141, Republic of Korea\\
\email{\{j.sun19,b.bhattarai,tk.kim\}@imperial.ac.uk}\\
\url{https://labicvl.github.io/}}

\maketitle

\begin{abstract}
We present a novel self-supervised learning approach for conditional generative adversarial networks (GANs) under a semi-supervised setting. Unlike prior self-supervised approaches which often involve geometric augmentations on the image space such as predicting rotation angles, our pretext task leverages the label space. We perform augmentation by randomly sampling sensible labels from the label space of the few labelled examples available and assigning them as target labels to the abundant unlabelled examples from the same distribution as that of the labelled ones. The images are then translated and grouped into positive and negative pairs by their target labels, acting as training examples for our pretext task which involves optimising an auxiliary match loss on the discriminator's side. We tested our method on two challenging benchmarks, CelebA and RaFD, and evaluated the results using standard metrics including Fr\'{e}chet Inception Distance, Inception Score, and Attribute Classification Rate. Extensive empirical evaluation demonstrates the effectiveness of our proposed method over competitive baselines and existing arts. In particular, our method surpasses the baseline with only 20\% of the labelled examples used to train the baseline.
\end{abstract}
\section{Introduction}
\label{sec:intro}
Face attribute and expression editing \cite{choi2018stargan,chen2016infogan,pumarola2018ganimation,karras2019style} has attracted tremendous attention thanks to the ongoing advancements in GANs \cite{goodfellow2014generative}, in particular conditional GANs (cGANs) \cite{odena2017conditional,mirza2014conditional,choi2018stargan,isola2017image,bhattarai2020inducing} which provide greater flexibility and control by incorporating labels in the generation process. However, deploying such cGANs in practice can be challenging as they rely heavily on large numbers of annotated examples. For instance, commonly used labelled datasets for training conditional GANs \cite{choi2018stargan,liu2019stgan,he2019attgan,mirza2014conditional} such as CelebA and ImageNet contain examples in the order of $10^5$ to $10^6$, which might be expensive to obtain in many applications.

To reduce the need of such huge labelled datasets in training cGANs, a promising approach is to utilise self-supervised methods which are successfully employed in a wide range of computer vision tasks including image classification \cite{gidaris2018unsupervised,zhai2019s}, semantic segmentation \cite{zhan2018mix}, robotics \cite{jang2018grasp2vec}, and many more. Recently, self-supervised learning is also gaining traction with GAN training \cite{luvcic2019high,chen2019self,tran2019self}, but prior work in this area \cite{luvcic2019high,chen2019self} has mostly focused on the input image space when designing the pretext task. For instance, \cite{chen2019self} proposed rotating images and minimising an auxiliary rotation loss similar to that of RotNet \cite{gidaris2018unsupervised}, which \cite{luvcic2019high} also adopted but in a semi-supervised setting. In general, existing methods mostly incorporate geometric augmentations on the input \emph{image space} as part of the pretext task. A main limitation of such approaches is their inability to generate new examples under each class label, for example rotating an image does not change its class label. In addition, our downstream task, attribute/expression editing, is more fine-grained in nature in comparison to tasks such as image classification on ImageNet. Therefore, we present a self-supervised method for training cGANs by making use of the \emph{label space} from the target domain and, if available, source domain as well. Specifically, under a semi-supervised setting wherein only few labelled examples are available, we utilise the large number of unlabelled examples to automatically generate additional labelled examples for our pretext task. Hence, our approach is orthogonal to existing methods.

\begin{figure}[t]
    \centering
    \includegraphics[width=0.95\linewidth]{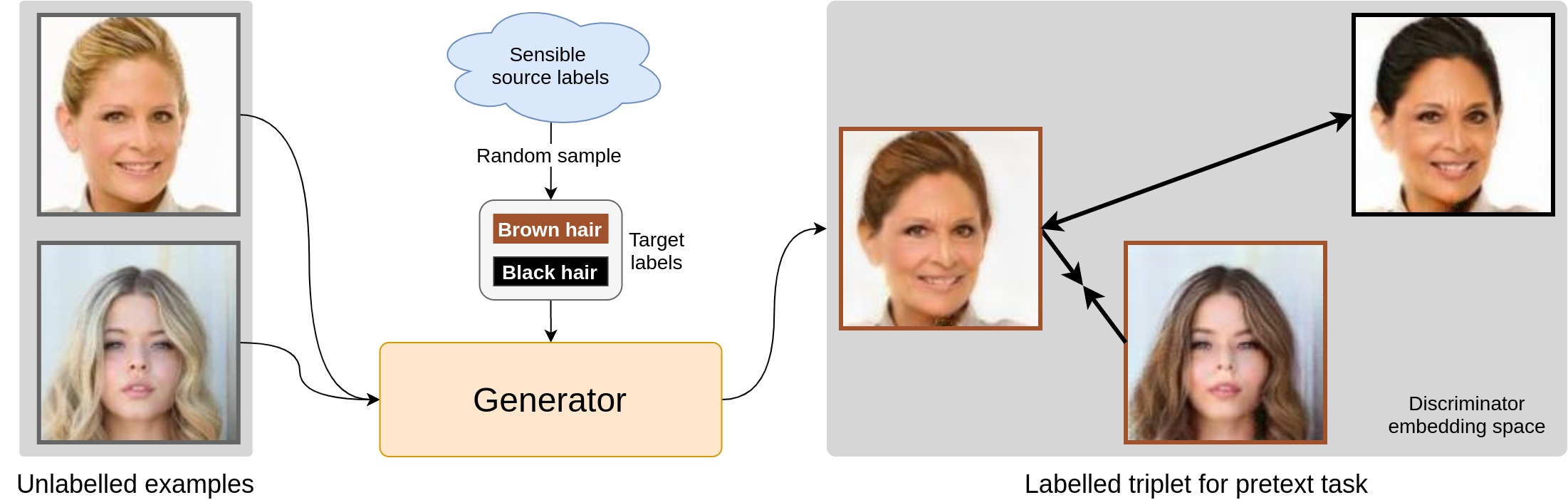}
    \caption{The procedure of generating triplet examples for our pretext task.}
    \label{fig:idea}
\end{figure}

Our idea draws inspirations from \cite{nair2018visual}, a self-supervised approach for reinforcement learning which trains a policy by randomly sampling imagined goals using a variational auto-encoder (VAE) \cite{kingma2014auto}. In a similar fashion, we can task the generator in a cGAN with synthesising images conditioned on randomly sampled target labels as a means to automatically provide additional supervision to the network. Motivated by our end goal of attribute editing and classification, we require that given unlabelled examples from the same distribution as the labelled ones, irrespective of their true source attribute labels, the generator should map the source images to similar regions of the synthetic image manifold if assigned the same target label and different regions otherwise. We treat every augmented target label vector as a unique state that needs to be reached on the translated domain regardless of the source labels of input images. In other words, whilst standard cGANs such as StarGAN \cite{choi2018stargan} and STGAN \cite{liu2019stgan} consider each component of the attribute vectors individually, our pretext task considers these vectors holistically.

% In particular, as facial attributes are multi-labelled and some attributes often co-occur (e.g. having blonde hair and being female are positively correlated in CelebA), such a constraint of matching attributes at a group level enables hierarchical clustering of attributes by mapping similar and coexisting attributes to the same region of the image manifold. As we only utilise sensible facial attribute labels (e.g. requiring that black hair and blonde hair do not coexist), our pretext task implicitly imposes such high-order behaviours of attributes.

Specifically, we propose to create a large pool of labelled examples by uniformly sampling labels from the source domain and assigning them to unlabelled data as their target labels. These unlabelled real images are then translated by the generator to create triplets of synthetic images as additional training examples for the generator (as illustrated in Figure \ref{fig:idea}). In addition, we also create such triplets using real images and their source labels from the small labelled pool to train the discriminator. Whilst triplets of real examples can help distill knowledge from the discriminator to the generator, both synthetic and real triplets are needed to maximise the benefits of our pretext task. The pretext task itself is trained using an auxiliary match loss optimised alongside existing losses of the baseline network \cite{caruana1997multitask}. This objective alleviates the overfitting problem for the discriminator in a semi-supervised setting as these triplets serve as additional supervision for the network. Unlike the standard triplet loss \cite{Weinberger2006distance,schroff2015facenet} which uses the Euclidean distance for comparison and fully shared weights between embeddings, we employ a learned convolutional head in the discriminator which takes \emph{concatenated pairs} of embeddings for comparison in a manner akin to learning a custom metric. However, instead of learning a metric, we employ a cross-entropy loss to directly classify pairs with matched labels and those with mismatched ones. Compared to linear loss functions such as the hinge loss, the cross-entropy loss allows for more precise probability estimations and ultimately better performance. Unlike \cite{chen2019self} which is purely geometric in nature, we view our approach as an image operation guided by augmented label codes performed on source images and is more in line with our end goal of attribute/expression editing.

We evaluated our method on two challenging benchmarks, CelebA and RaFD, which are popular benchmarks for facial attribute and expression translations. We take StarGAN~\cite{choi2018stargan} as our baseline cGAN, but our method is generic in nature. We compared the results both quantitatively and qualitatively. We used standard metrics Fr\'{e}chet Inception Distance (FID) and Inception Score (IS) for quantitative comparisons.

\section{Related Work}
\label{label:related_works}
\noindent \textbf{Image-to-Image (I2I) Translation with cGANs.} cGANs \cite{mirza2014conditional} incorporate labels as additional inputs, allowing the network to handle multiple modalities and providing greater flexibility and control over generated examples. I2I translation and facial attribute editing frameworks such as Pix2pix \cite{isola2017image} and IcGAN \cite{perarnau2016icgan} have greatly benefited from employing cGANs, with IcGAN allowing for multi-attribute manipulation without needing to be retrained for different source-target combinations. StarGAN \cite{choi2018stargan} and AttGAN \cite{he2019attgan} both improve upon IcGAN by using an end-to-end framework, an encoder-decoder architecture for the generator, and a cycle-consistency loss. STGAN \cite{liu2019stgan} further improves upon these frameworks by using the difference between source and target labels as conditional input to the generator. All these methods rely heavily on source attribute labels which can be difficult to obtain in practical applications.

\noindent \textbf{Self-Supervised Learning.} Self-supervised methods have been successfully employed to fill the gap between unsupervised and supervised frameworks, particularly for image classification tasks. Well-known self-supervised approaches include predicting relative positioning of image patches \cite{doersch2015unsupervised}, generating image content from surroundings \cite{pathak2016context}, colouring greyscale images \cite{zhang2016colorful}, counting visual primitives \cite{noroozi2017representation}, and predicting rotation angles \cite{gidaris2018unsupervised}. These pretext tasks all involve certain artificially designed geometric transformations on the input images. However, it might be challenging to choose the transformation most optimal for a specific task. Recently, a few approaches have been proposed which rely purely on the model's interaction with data, particularly in reinforcement learning. Grasp2Vec \cite{jang2018grasp2vec} learns object-centric visual embeddings purely through autonomous interaction between a robot and the environment. \cite{nair2018visual} uses a VAE to randomly sample imagined goals for the agent to perform. Both these frameworks serve as inspirations for MatchGAN.

\noindent \textbf{Self- and Semi-Supervised Learning in GANs.} Semi-supervised learning methods become relevant in situations where there are limited number of labelled examples and a large number of unlabelled ones. One of the popular approaches is to annotate unlabelled data with pseudo-labels \cite{Lee2013PseudoLabelT}. Self-supervised approaches have also been explored in semi-supervised learning settings. For example, \cite{zhai2019s} employed the rotation loss \cite{gidaris2018unsupervised} and outperformed fully-supervised methods with a fraction of examples labelled. As for GANs, \cite{chen2019self} proposed to minimise the rotation loss \cite{gidaris2018unsupervised} on the discriminator, mitigating the discriminator-forgetting problem and allowing more stable representations to be learned. In a semi-supervised setting, \cite{luvcic2019high} proposed training an auxiliary classifier with the few labelled data which is then used to annotate the unlabelled data with pseudo-labels, and \cite{yaxing2020semi} differs from \cite{luvcic2019high} by adding these pseudo-labels progressively and through consensus. These method, however, are reliant on the performance of the auxiliary classifier and add significant complexity to the training process.
%cannot be used for cross-domain training, whereas MatchGAN is agnostic to the source domain information.

\section{Method}
\label{sec:method}
Our task is to perform I2I translation in a semi-supervised setting where the \textit{majority} of training examples are unlabelled except for a \textit{small} number. As training a large network in such a scenario could lead to overfitting, we aim to mitigate the problem by providing weak supervision using the large number of unlabelled examples available. In short, we propose to utilise the translated images and their associated target labels as extra training examples for a pretext task. The goal of the pretext task is to minimise an auxiliary match loss classifying positive and negative pairs of images in a manner akin to metric learning. Compared to optimising a cross-entropy loss across all possible target labels, this approach is more efficient and has been successfully adopted in one-shot learning \cite{koch2015siamese} and face recognition \cite{schroff2015facenet}. We use StarGAN \cite{choi2018stargan} as the baseline for our experiments, and as a result we will give a brief overview of its architecture and loss functions before introducing our method. However, we emphasise that our method is generic in nature and can be applied to any other cGAN.
 
\subsection{Background on StarGAN}
\noindent \textbf{Overview.} Here we provide a brief background on cGANs taking reference from StarGAN \cite{choi2018stargan} but in a semi-supervised setting. Let $X$ be the set of source images and $Y$ the labels, where $X$ is partitioned into labelled and unlabelled subsets, $X^L$ and $X^U$, respectively. StarGAN aimed to tackle the problem of multi-domain I2I translation without having to train a new GAN for each domain pair. It accomplished this by encoding target domain information as binary or one-hot labels and feeding them along with source images to the generator. During training, the generator $G$ is required to translate a source image $x\sim X$ conditioned on a target domain label $y\sim Y$. The discriminator $D$ receives an image and produces an embedding $D_{emb}(x)$, which is then used to produce two outputs $D_{adv}(x)$ and $D_{cls}(x)$. The former, $D_{adv}(x)$, is used to optimise the Wasserstein GAN with gradient penalty \cite{ishaan2017improved} defined by
\begin{equation}
    \mathcal{L}_{adv} = \mathop{\mathbb{E}}_{x\sim X;y\sim Y}[D_{adv}(G(x,y))] - \mathop{\mathbb{E}}_{x\sim X}[D_{adv}(x)]  + \lambda_{gp}\mathop{\mathbb{E}}_{\hat{x}\in \hat{X}}[\|\nabla_{\hat{x}}D(\hat{x})\|_2-1]^2,
\end{equation}
where $\hat{X}$ consists of points uniformly sampled from straight lines between $X$ and the synthetic distribution $G(X,Y)$.
The latter output $D_{cls}(x)$ consists of probabilities over attributes/expressions used for optimising a classification loss to help guide $G$ towards generating images that more closely resemble the target domain. The classification loss for $D$ and $G$ are given by
\begin{align}
    \mathcal{L}_{cls}^D &= \mathop{\mathbb{E}}_{x\sim X^L_y;y\sim Y}[-y\cdot\log(D_{cls}(x))],\\
    \mathcal{L}_{cls}^G &= \mathop{\mathbb{E}}_{x\sim (X^L\cup X^U);y\sim Y}[-y\cdot\log(D_{cls}(G(x,y)))]
\end{align}
respectively, where the subset $X^L_y\subset X^L$ consists of examples with label $y$. In addition, a cycle-consistency loss \cite{CycleGAN2017},
\begin{equation}
    \mathcal{L}_{cyc} = \mathop{\mathbb{E}}_{x\sim X^L_y;y,y'\sim Y} [\|x-G(G(x,y'),y)\|_1],
\end{equation}
is incorporated to ensure that $G$ preserves content unrelated to the domain translation task. The overall objective for StarGAN is given by
\begin{align}
    \mathcal{L}_D = \mathcal{L}_{adv} + \lambda_{cls}\mathcal{L}_{cls}^D,\qquad \mathcal{L}_G = -\mathcal{L}_{adv} + \lambda_{cls}\mathcal{L}_{cls}^G + \lambda_{cyc}\mathcal{L}_{cyc}.
\end{align}
\textbf{Achitectural Details.} StarGAN is fully convolutional. Its generator consists of 3 downsampling convolutional layers, 6 bottleneck residual blocks, and 3 upsampling convolutional layers. Each downsampling or upsampling layer halves or doubles the spatial dimensions of the input. Instance normalisation and ReLU is used for all layers except the output layer. The discriminator consists of 6 downsampling convolutional layers with leaky ReLUs with a slope of 0.01 for negative values. The discriminator output $D_{emb}(x)$ has $2048$ channels and is then fed through two separate convolutional heads to produce $D_{adv}(x)$ and $D_{cls}(x)$, with $D_{cls}(x)$ having passed through an additional Softmax layer if ground truths are one-hot or Sigmoid layer otherwise. With an input image size of $128\times 128$, StarGAN has 53.22M learned parameters in total, comprising 8.43M from the generator and 44.79M from the discriminator.

\subsection{Triplet Matching Objective as Pretext Task}
\begin{figure}[t]
  \centering
  \includegraphics[width=0.99\linewidth]{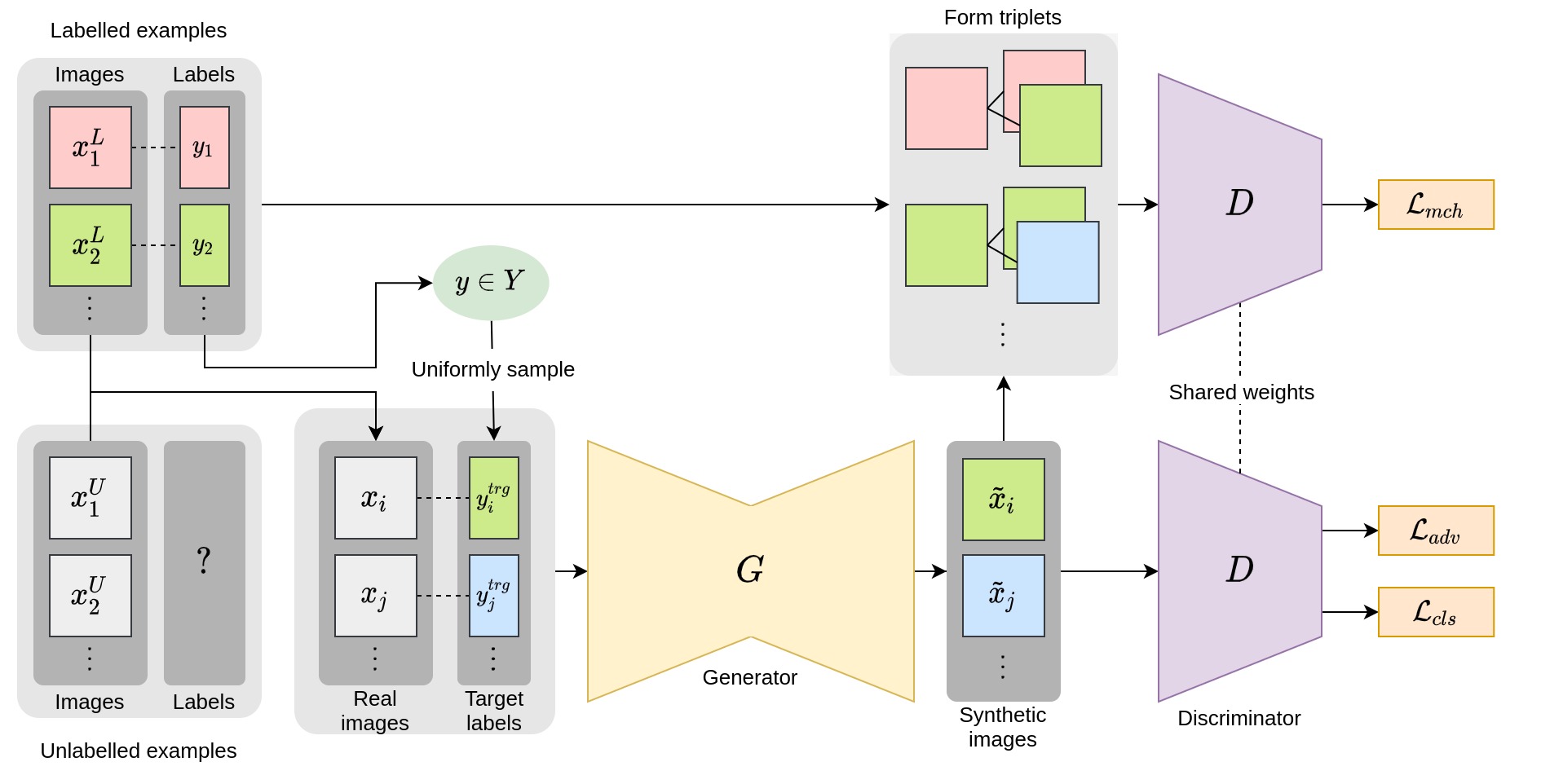}
  \caption{Detailed pipeline of the proposed framework. Our contribution lies in augmenting the label space of the small-scale labelled pool followed by assigning them to the large-scale unlabelled pool as their target labels. We then generate triplets containing both matched and mismatched pairs based on the target labels in the synthetic domain and source labels in the real domain. Finally, we minimise a match loss as an auxiliary loss to the existing framework.}
  \label{fig:pipeline}
\end{figure}

\noindent \textbf{Pretext from Synthetic Data.} Whilst existing self-supervised approaches mostly rely on geometric transformation of input images, we take the inspiration of utilising target domain information from reinforcement learning literature. \cite{jang2018grasp2vec} learned an embedding of object-centric images by comparing the difference prior to and after an object is grasped. However, this requires source information which can be scarce in semi-supervised learning settings. \cite{nair2018visual} utilised a variational autoencoder \cite{kingma2014auto} to randomly generate a large amount of goals to train the agent in a self-supervised manner. Similar to \cite{nair2018visual}, our self-supervised method involves translating unlabelled images to random target domains and using the resulting synthetic images to optimise a match loss (see Figure \ref{fig:pipeline}).

Recently \cite{chen2019self} proposed to minimize the rotation loss \cite{gidaris2018unsupervised} on the discriminator to mitigate its forgetting problem due to the continuously changing generator distribution. Compared to this work which involves only four rotations, the number of possible goals in our setting grows exponentially with respect to the number of attributes (CelebA is multi-labelled) and this would be challenging to implement with a softmax in the same way. Imposing a triplet-like constraint also forces the generator to maintain consistency on translated attributes, ultimately allowing attributes to be retained better on synthetic images. Hence, we propose an auxiliary match loss based on label information as a pretext task for both $G$ and $D$. A triplet consists of an anchor example $x_a$, a positive example $x_p$ which shares the same label information as $x_a$, and a negative example $x_n$ which has a different label. Unlike the standard triplet loss \cite{Weinberger2006distance,schroff2015facenet}, we concatenate the discriminator embeddings of the positive pair $(D_{emb}(x_a), D_{emb}(x_p))$ and negative pair $(D_{emb}(x_a), D_{emb}(x_n))$ respectively along the channel axis and feed them through a single convolutional layer, producing probability distributions $D_{mch}(x_a, x_p)$ and $D_{mch}(x_a, x_n)$ respectively over whether each pair has matching labels. Specifically, we propose the following triplet matching objective
\begin{flalign}\label{eqn:triplet}
    &&\mathcal{L}_{mch}^{D} &= \mathop{\mathbb{E}}_{\substack{x_a,x_p\sim X^L_y;x_n\sim X^L_{y'}\\y\neq y'\sim Y}} -[\log(D_{mch}(x_a, x_p)) + \log(1-D_{mch}(x_a, x_n))],
\end{flalign}
\begin{flalign}
    \mathcal{L}_{mch}^{G} = \mathop{\mathbb{E}}_{x_1,x_2,x_3\sim (X^L\cup X^U); y\neq y'\sim Y} &-[\log(D_{mch}(G(x_1,y), G(x_2,y))) \notag \\
    & +\log(1-D_{mch}(G(x_1,y), G(x_3,y')))].
\end{flalign}
Rather than using the standard triplet loss which sticks with the Euclidean distance as a single measurement, concatenation allows the network to continuously adapt itself to the pattern in the data and thus acquire more optimal ways of carrying out such comparisons, in a manner similar to learning a custom metric\footnote{However, concatenation does not enforce symmetry so it is not strictly a metric.}. In addition, the cross-entropy loss allows the network to make more precise probability estimations compared to linear loss functions and ultimately learn more refined representations. Our overall loss function is given by
\begin{align}
    \mathcal{L}_D &= \mathcal{L}_{adv} + \lambda_{cls}\mathcal{L}_{cls}^D + \lambda_{mch}\mathcal{L}_{mch}^{D}, \label{eqn:D_loss} \\
    \mathcal{L}_G &= -\mathcal{L}_{adv} + \lambda_{cls}\mathcal{L}_{cls}^G + \lambda_{cyc}\mathcal{L}_{cyc} + \lambda_{mch}\mathcal{L}_{mch}^{G}. \label{eqn:G_loss}
\end{align}
As some of the components in Equation \ref{eqn:D_loss} and \ref{eqn:G_loss} require source labels, they cannot be directly implemented on unlabelled examples in a semi-supervised setting. As a result, we train the network with labelled and unlabelled examples in an alternating fashion as detailed in Algorithm \ref{alg:training}.

\begin{algorithm}[!t]
\caption{MatchGAN.}
\label{alg:training}
	\begin{algorithmic}[1]
	    \State \textbf{Input:} Labelled set $X^L$ with set of all possible class labels $Y=\{y_1,\ldots,y_K\}$ separated into disjoint sets $X^L=X_1^L\sqcup \cdots \sqcup X_K^L$ by label, and unlabelled set $X^U$.
	    \State \textbf{Initialise:} Generator $G$, Discriminator $D$, weights $\theta_G$ and $\theta_D$, learning rates $\eta_G$ and $\eta_D$, \# of iterations $N$, batch size $B$, \# of $D$ updates per $G$ update $n_G$.
		\For {$i = 1,\ldots,N$}
			\If{$i$ is odd}
			    \parState{Form a batch of $B$ real images and labels $(R^{(i)}, Y_{src}^{(i)})$ chosen uniformly from $k$ classes $\mathcal{K}_{src}\subset \{1,\ldots,K\}$, where $R^{(i)}=\bigsqcup_{j\in \mathcal{K}_{src}}R_{j}^{(i)}$ and each $R_j^{(i)}\subset X_j^L$.}
    			\parState{$\mathcal{L}_{cls}^D \leftarrow \frac{1}{B}\sum_{(r,y)\in (R^{(i)},Y_{src}^{(i)})}-y\cdot\log(D_{cls}(r))$.}
    			\parState{Get $T_R^{(i)}=\{(x_a,x_p,x_n):x_a,x_p\in R_{k_1}^{(i)}\text{ and }x_n\in R_{k_2}^{(i)}, k_1\neq k_2\in\mathcal{K}_{src}\}$, a set of triplets sampled from the mini-batch $R^{(i)}$.}
    			\parState{$\displaystyle \mathcal{L}_{mch}^{D} \leftarrow \frac{1}{|T_R^{(i)}|}\sum_{(x_a,x_p,x_n)\in T_R^{(i)}} -[\log(D_{mch}(x_a, x_p)) + \log(1-D_{mch}(x_a, x_n))]$.}
			\Else
			    \parState{Sample mini-batch of $B$ unlabelled real images $R^{(i)}\subset X^U$.}
    		\EndIf
    		\parState{Form a batch of $B$ target labels $Y_{trg}^{(i)}$ chosen uniformly from $k$ classes $\mathcal{K}_{trg}\subset \{1,\ldots,K\}$.}
    		\parState{Generate fake images $F^{(i)}=\{G(r,y):(r,y)\in(R^{(i)},Y_{trg}^{(i)})\}$.}
			\parState{$\mathcal{L}_{adv}^D \leftarrow \frac{1}{B}\sum_{(r,f)\in (R^{(i)},F^{(i)})}[D_{adv}(f) - D_{adv}(r)  + \lambda_{gp}(\|\nabla_{\hat{x}}D_{adv}(\hat{x})\|_2-1)^2]$, where $\hat{x}=\alpha r + (1-\alpha)f$ and $\alpha\sim U(0,1)$ is random.}
		    \parState{$\theta_D \leftarrow Adam\left(\nabla_{\theta_D} (\mathcal{L}_{adv}^D + \text{odd}(i)(\lambda_{cls}\mathcal{L}_{cls}^D + \lambda_{mch}\mathcal{L}_{mch}^{D})),\eta_D\right)$ using Adam \cite{Kingma2015Adam}, where $odd(i)=1$ if $i$ is odd or $0$ otherwise.}
		    \If{$i$ is a multiple of $n_G$}
		        \If{$i$ is odd}
    		        \parState{$\mathcal{L}_{cyc} \leftarrow \frac{1}{B}\sum_{(r,y,y')\in(R^{(i)},Y_{src}^{(i)},Y_{trg}^{(i)})} \|r-G(G(r,y'),y)\|_1$.}
    		    \EndIf
    		    \parState{$\mathcal{L}_{adv}^G \leftarrow \frac{1}{B}\sum_{f\in F^{(i)}}-D_{adv}(f)$.}
		        \parState{$\mathcal{L}_{cls}^G \leftarrow \frac{1}{B}\sum_{(f,y)\in (F^{(i)},Y_{trg}^{(i)})} -y\cdot\log(D_{cls}(G(f,y)))$.}
		        \parState{Get $T_F^{(i)}=\{(x_a,x_p,x_n):x_a,x_p\in F_{k_1}^{(i)}\text{ and }x_n\in F_{k_2}^{(i)}, k_1\neq k_2\in\mathcal{K}_{trg}\}$, a set of triplets sampled from the mini-batch $F^{(i)}$, where $F^{(i)} = \bigsqcup_{j\in \mathcal{K}_{trg}} F_{j}^{(i)}$ and each $F_{j}^{(i)}$ corresponds to target label $y_j$.}
		        \parState{$\displaystyle \mathcal{L}_{mch}^{G} \leftarrow \frac{1}{|T_F^{(i)}|}\sum_{(x_a,x_p,x_n)\in T_F^{(i)}} -[\log(D_{mch}(x_a, x_p)) + \log(1-D_{mch}(x_a, x_n))]$.}
		        \parState{$\theta_G \leftarrow Adam\left(\nabla_{\theta_G} (\mathcal{L}_{adv} + \lambda_{cls}\mathcal{L}_{cls}^G + \lambda_{mch}\mathcal{L}_{mch}^{G} + \text{odd}(i)\lambda_{cyc}\mathcal{L}_{cyc}), \eta_G\right)$.}
		    \EndIf
		\EndFor
		\State \textbf{Output:} Optimal $G$.
	\end{algorithmic}
\end{algorithm}

\begin{figure}[!t]
    \centering
    \includegraphics[width=0.9\linewidth]{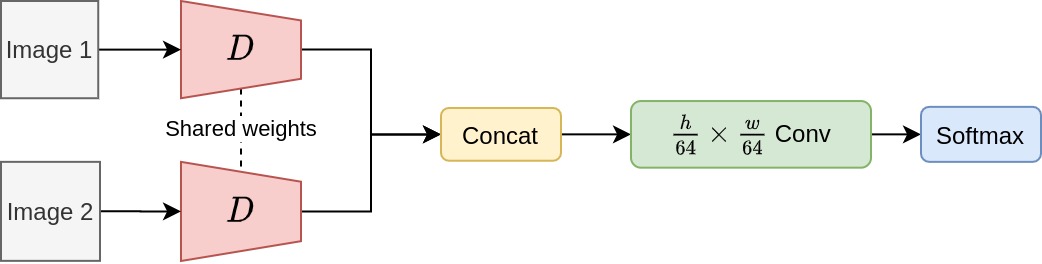}
    \caption{Architecture of the match loss head, where $h$ and $w$ are the height and width of the input images.}
    \label{fig:match_loss}
\end{figure}

\noindent\textbf{Architectural Details.} MatchGAN is built directly on top of StarGAN but includes an additional head for $\mathcal{L}_{mch}$ after the discriminator output $D_{emb}(x)$ (see Figure \ref{fig:match_loss}). Specifically, a triplet of images $(x_a, x_p, x_n)$ are passed through $D$ to produce embeddings $(D_{emb}(x_a),D_{emb}(x_p),D_{emb}(x_n))$. The positive and negative pairs $(D_{emb}(x_a),D_{emb}(x_p))$ and $(D_{emb}(x_a),D_{emb}(x_n))$ are concatenated respectively along the channel dimension to produce 4096-channel embeddings. These embeddings are then convolved and passed through a Softmax layer to produce probabilities of whether each image pair is matched. For input images of size $128\times 128$, this head adds approximately 32.77K to the total number of learned parameters which is negligible compared to the 53.22M parameters in the StarGAN baseline, and thus has very little impact on training efficiency.

\section{Experiments}
\label{sec:experiments}
\subsection{Implementation details}
We used StarGAN \cite{choi2018stargan} as a baseline for our experiments\footnote{Code and pretrained model at \url{https://github.com/justin941208/MatchGAN}.}. StarGAN unifies multi-domain image-to-image translation with a single generative network and is well suited to our label-based self-supervised approach. However, we would like to re-emphasise that our method is a general idea and can be extended to other cGANs. To avoid potential issues during training, we used the same hyperparameters as the original StarGAN. Specifically, we trained the network for 200K discriminator iterations with 1 generator update after every 5 discriminator updates. We used the Adam optimiser \cite{Kingma2015Adam} with $\beta_1=0.5$ and $\beta_2=0.999$, and the initial learning rates for both generator and discriminator were set to $10^{-4}$ for the first 100K iterations and decayed linearly to 0 for the next 100K. We trained the model using mini-batches of 16 examples (sampled from 4 random classes if from the labelled pool) and mapped to 4 random target classes. Training took approximately 10 hours to complete on an NVIDIA RTX 2080Ti GPU.

\subsection{Datasets}
We evaluated our method on two challenging face attributes and expression manipulation datasets, The CelebFaces Attributes Dataset (CelebA) \cite{liu2015faceattributes} and The Radboud Faces Database (RaFD) \cite{Langner2010Presentation}. Both datasets were split into training and test sets, and we report results on the test set.

\noindent\textbf{CelebA.} CelebA contains 202,599 images of celebrities of size $178\times 218$ with 40 attribute annotations. We selected 5 attributes including 3 hair colours (black, blond, and brown), gender, and age. The images were cropped to $178\times 178$ then resized to $128\times 128$. The experiments followed the official partition of 162,770 examples for training and 19,962 for testing. We created a semi-supervised scenario with limited labelled training examples by uniformly sub-sampling a percentage of training examples as labelled and setting the rest as unlabelled. The sub-sampling process was done to ensure that the examples were spread evenly between classes whenever possible to avoid potential class imbalance issues.

\noindent\textbf{RaFD.} RaFD is a much smaller dataset with 8,040 images of size $681\times 1024$ of 67 identities of different genders, races, and ages displaying 8 emotional expressions. The images were cropped to $600\times 600$ (centred on face) before being resized to $128\times 128$. A total of 7 randomly selected identities, comprising 840 images, were chosen as the test set and the rest (60 identities comprising 7200 images) as the training set. Similar to CelebA, a semi-supervised setting was created by splitting the training set into labelled and unlabelled pools.

\subsection{Baseline}
Our baseline was established by setting $\lambda_{mch}$ to 0 whilst leaving all other procedures unchanged. As for MatchGAN, the value of $\lambda_{mch}=0.5$ was used for all experiments. To verify that our method is scalable to both small and large number of annotated data, we tested our approach with various percentages of training examples labelled. Specifically, we performed experiments setting 1\%, 5\%, 10\%, and 20\% of CelebA training data as labelled examples, and similarly for 10\%, 20\%, and 50\% of RaFD training data as RaFD is a significantly smaller dataset. Finally, we also evaluated our method on the full datasets to verify the effectiveness of our method on benchmarks designed for supervised learning. We also tested the rotation loss \cite{chen2019self} for comparison.

\subsection{Evaluation metrics}
We employed the Fr\'{e}chet Inception Distance (FID) \cite{heusel2017gans} and Inception Score (IS) \cite{salimans2016improved} for quantitative evaluations. FID measures the similarity between the distribution of real examples and that of the generated ones by comparing their respective embeddings from a pretrained Inception-v3 network \cite{Szegedy2016Rethinking}. IS also measures image quality but relies on the probability outputs of the same Inception-v3 network, taking into account the meaningfulness of individual images and the diversity of all images. If the generated images are of high quality, then FID should be small whereas IS should be large. We computed the FID by translating test images to each of the target attribute domains (5 for CelebA, 8 for RaFD) and comparing the distributions before and after translations. The IS was computed as an average obtained from a 10-fold split of the test set.

In addition to FID and IS, we also used GAN-train and GAN-test \cite{shmelkov2018good} to measure the attribute classification rate of translated images.
In short, given a set of real images $X$ with a train-test split $X=X_{train}\sqcup X_{test}$, GAN-train is the accuracy obtained from a classifier trained on synthetic images $G(X_{train})$ and tested on real images $X_{test}$, whereas for GAN-test the classifier is trained on real images $X_{train}$ and tested on synthetic images $G(X_{test})$.

\subsection{Ablation studies}
MatchGAN involves extracting triplets from labelled real examples and all synthetic examples - labelled and unlabelled. To show that the proposed method does not simply rely on the few labelled examples and that both unlabelled and synthetic examples are necessary to achieve good performance, the network was trained in several other scenarios in which various amounts of real and synthetic data used for updating the match loss $\mathcal{L}_{mch}$ were removed (shown in Table \ref{tab:ablation}). A few observations can be made from this table. First, including a large number of unlabelled data is essential for improving performance, which is clear from comparing A and B with the rest. Second, incorporating the match loss $\mathcal{L}_{mch}$ provides substantial improvement in performance, as observed from A vs B, and C vs D--G. This improvement was achieved despite the match loss not utilising all the available data, as seen from setups D--F. Third, match loss indeed benefits from training with synthetic examples which is evident from C vs D, and E vs F--G. Fourth, unlabelled synthetic examples can be used to achieve further performance improvement, as seen from F vs G. In addition, H was trained using the standard triplet loss \cite{schroff2015facenet} which G also outperforms. Therefore, G will be used as the default setup for MatchGAN in all following experiments.
\begin{table}
    \centering
    \caption{The results of the ablation studies -- FID scores obtained using various amounts of training data. Setups A and C are baseline StarGAN \cite{choi2018stargan}, whereas the other setups update the match loss $\mathcal{L}_{mch}$ using different portions of the training data.}
    \begin{tabularx}{\linewidth}{Y|c|c|c|c|c|c|c|c|c}
        \thickhline
        \multicolumn{2}{c|}{Setup} & A & B & C & D & E & F & G & H\cite{schroff2015facenet} \\ \hline
        \multicolumn{2}{c|}{Total number of training examples} & 2.5K & 2.5K & 162K & 162K & 162K & 162K & 162K & 162K \\ \hline
        Number of & Real (labelled) & 0 & 2.5K & 0 & 0 & 2.5K & 2.5K & 2.5K & 2.5K \\
        examples for & Synthetic (labelled) & 0 & 2.5K & 0 & 2.5K & 0 & 2.5K & 2.5K & 2.5K \\
        $\mathcal{L}_{mch}$ & Synthetic (unlabelled) & 0 & 0 & 0 & 160K & 0 & 0 & 160K & 160K \\ \hline
        \multicolumn{2}{c|}{FID$\downarrow$} & 24.20 & 17.26 & 16.11 & 13.78 & 13.66 & 10.88 & \textbf{9.43} & 14.86\\
        % \multicolumn{2}{c|}{IS$\uparrow$} & 2.85 & 2.91 & 2.89 & 2.89 & 2.94 & \textbf{2.96} & 2.94 \\
        \thickhline
    \end{tabularx}
    \label{tab:ablation}
\end{table}

\subsection{Quantitative evaluations}
We evaluated the performance of our proposed method, the baseline, and rotation loss \cite{chen2019self} using FID and IS and the results are shown in Table \ref{tab:fid_is}. In terms of FID, MatchGAN consistently outperformed the baseline in both CelebA and RaFD. For CelebA in particular, with just 20\% of training examples labelled, our method was able to achieve better performance than the baseline with 100\% of the training examples labelled. Our method also has a distinct lead over the baseline when there are very few labelled examples. In addition, our method was also on par with or even outperformed rotation loss in both datasets, again with a distinct advantage over rotation loss when labelled examples are limited.

In terms of IS, we still managed to outperform both the baseline and rotation loss in the majority of the setups. In other setups our method was either on par with the baseline or slightly underperforming within a margin of 0.02. We would like to emphasise that IS is less consistent than FID as it does not compare the synthetic distribution with an ``ideal'' one. In addition, IS is computed using the 1000-dimensional output of Inception-v3 pretrained on ImageNet which is arguably less suitable for human face datasets such as CelebA and RaFD. However, we included IS here as it is still one of the most widely used metrics for evaluating the performance of GANs.
\begin{table}
    \centering
    \caption{Baseline vs Rotation vs MatchGAN in terms of FID and IS scores.}
    \begin{tabularx}{\linewidth}{c|c|c|Y|Y|Y|Y|Y||Y}
        \thickhline
        \multirow{2}{*}{Dataset} & \multirow{2}{*}{Metric} & \multirow{2}{*}{Setup} & \multicolumn{6}{c}{Percentage of training data labelled} \\ \cline{4-9}
        & & & 1\% & 5\% & 10\% & 20\% & 50\% & 100\% \\ \hline
        \multirow{6}{*}{CelebA} &\multirow{3}{*}{FID$\downarrow$} & Baseline~\cite{choi2018stargan} & 17.04 & 10.54 & 9.47 & 7.07 & \diagbox[dir=NE]{ }{ } & 6.65 \\ %\cline{3-9}
        & & Rotation~\cite{chen2019self} & 17.08 & 10.00 & \textbf{8.04} & 6.82 & \diagbox[dir=NE]{ }{ } & 5.91 \\
        & & MatchGAN & \textbf{12.31} & \textbf{9.34} & 8.81 & \textbf{6.34} & \diagbox[dir=NE]{ }{ } & \textbf{5.58} \\ \cline{2-9}
        & \multirow{3}{*}{IS$\uparrow$} & Baseline~\cite{choi2018stargan} & 2.86 & 2.95 & \textbf{3.00} & 3.01 & \diagbox[dir=NE]{ }{ } & 3.01\\ %\cline{3-9}
        & & Rotation~\cite{chen2019self} & 2.82 & \textbf{2.99} & 2.96 & 3.01 & \diagbox[dir=NE]{ }{ } & 3.06 \\
        & & MatchGAN & \textbf{2.95} & 2.95 & 2.99 & \textbf{3.03} & \diagbox[dir=NE]{ }{ } & \textbf{3.07} \\ \hline
        \multirow{6}{*}{RaFD} &\multirow{3}{*}{FID$\downarrow$} & Baseline~\cite{choi2018stargan} & \diagbox[dir=NE]{ }{ } & \diagbox[dir=NE]{ }{ } & 32.015 & 11.75 & 7.24 & 5.14 \\ %\cline{3-9}
        & & Rotation~\cite{chen2019self} & \diagbox[dir=NE]{ }{ } & \diagbox[dir=NE]{ }{ } & 28.88 & 10.96 & \textbf{6.57} & \textbf{5.00} \\
        & & MatchGAN & \diagbox[dir=NE]{ }{ } & \diagbox[dir=NE]{ }{ } & \textbf{22.75} & \textbf{9.94} & 6.65 & 5.06 \\ \cline{2-9}
        & \multirow{3}{*}{IS$\uparrow$} & Baseline~\cite{choi2018stargan} & \diagbox[dir=NE]{ }{ } & \diagbox[dir=NE]{ }{ } & \textbf{1.66} & 1.60 & 1.58 & 1.56 \\ %\cline{3-9}
        & & Rotation~\cite{chen2019self} & \diagbox[dir=NE]{ }{ } & \diagbox[dir=NE]{ }{ } & 1.62 & 1.58 & 1.58 & \textbf{1.60} \\
        & & MatchGAN & \diagbox[dir=NE]{ }{ } & \diagbox[dir=NE]{ }{ } & 1.64 & \textbf{1.61} & \textbf{1.59} & 1.58 \\
        \thickhline
    \end{tabularx}
    \label{tab:fid_is}
\end{table}

In terms of GAN-train and GAN-test classification rates, our method outperformed the baseline in both CelebA and RaFD (shown in Table \ref{tab:acr}) under the 100\% setup which has the best FID overall. MatchGAN again obtained a higher GAN-train accuracy than the baseline, indicating that the synthetic examples generated by MatchGAN can be more effectively used to augment small data for training classifiers. We report the results under the 100\% setup as it has the lowest FID and that FID is considered one of the most robust metrics for evaluating the performance of GANs. We expect GAN-train and GAN-test in other setups to be proportional to their respective FIDs as well.
\begin{table}
    \centering
    \caption{Baseline vs MatchGAN in terms of GAN-train and GAN-test classification rate under the 100\% setup. GAN-train for CelebA and GAN-test were obtained by averaging individual attribute accuracies, whereas top-1 accuracy was used when computing GAN-train for RaFD.}
    \begin{tabularx}{\linewidth}{Y|Y|Y|Y}
        \thickhline
        Dataset & Setup & GAN-train & GAN-test \\ \hline
        \multirow{2}{*}{CelebA} & Baseline \cite{choi2018stargan} & 87.29\% & 81.11\% \\ \cline{2-4}
        & MatchGAN & \textbf{87.43\%} & \textbf{82.26\%} \\ \hline
        \multirow{2}{*}{RaFD} & Baseline \cite{choi2018stargan} & 95.00\% & 75.00\% \\ \cline{2-4}
        & MatchGAN & \textbf{97.78\%} & \textbf{75.95\%} \\
        \thickhline
    \end{tabularx}
    \label{tab:acr}
\end{table}

\subsection{Qualitative evaluations}
\begin{figure}
    \centering
    \includegraphics[width=0.9\linewidth]{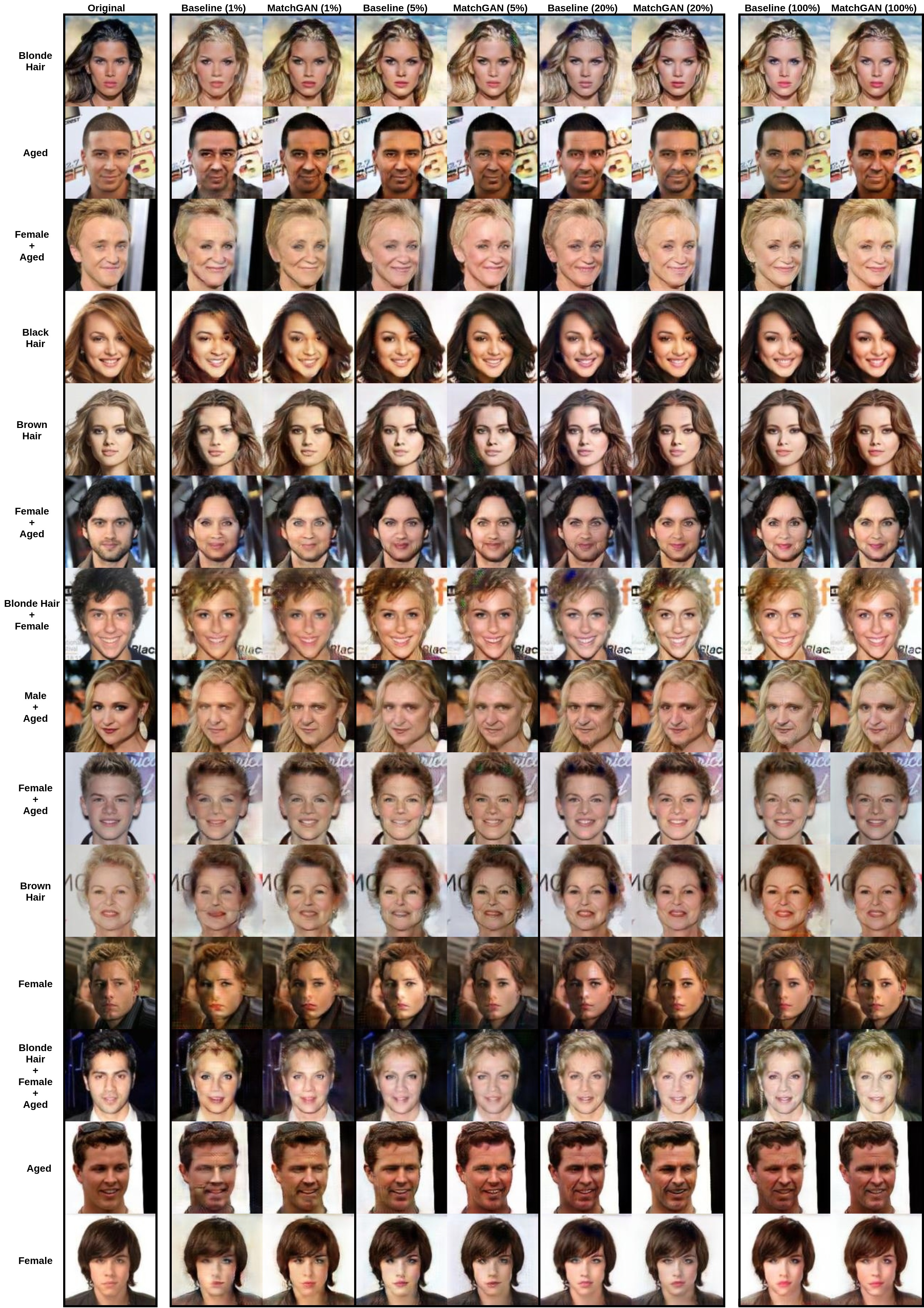}
    \caption{Synthetic examples of MatchGAN vs Baseline on CelebA (zoom in for a better view). Each row corresponds to a single- or multi-attribute manipulation, with target attributes listed on the left side.}
    \label{fig:qualitative_celeba}
\end{figure}

\begin{figure}[!t]
    \centering
    \includegraphics[width=0.9\linewidth]{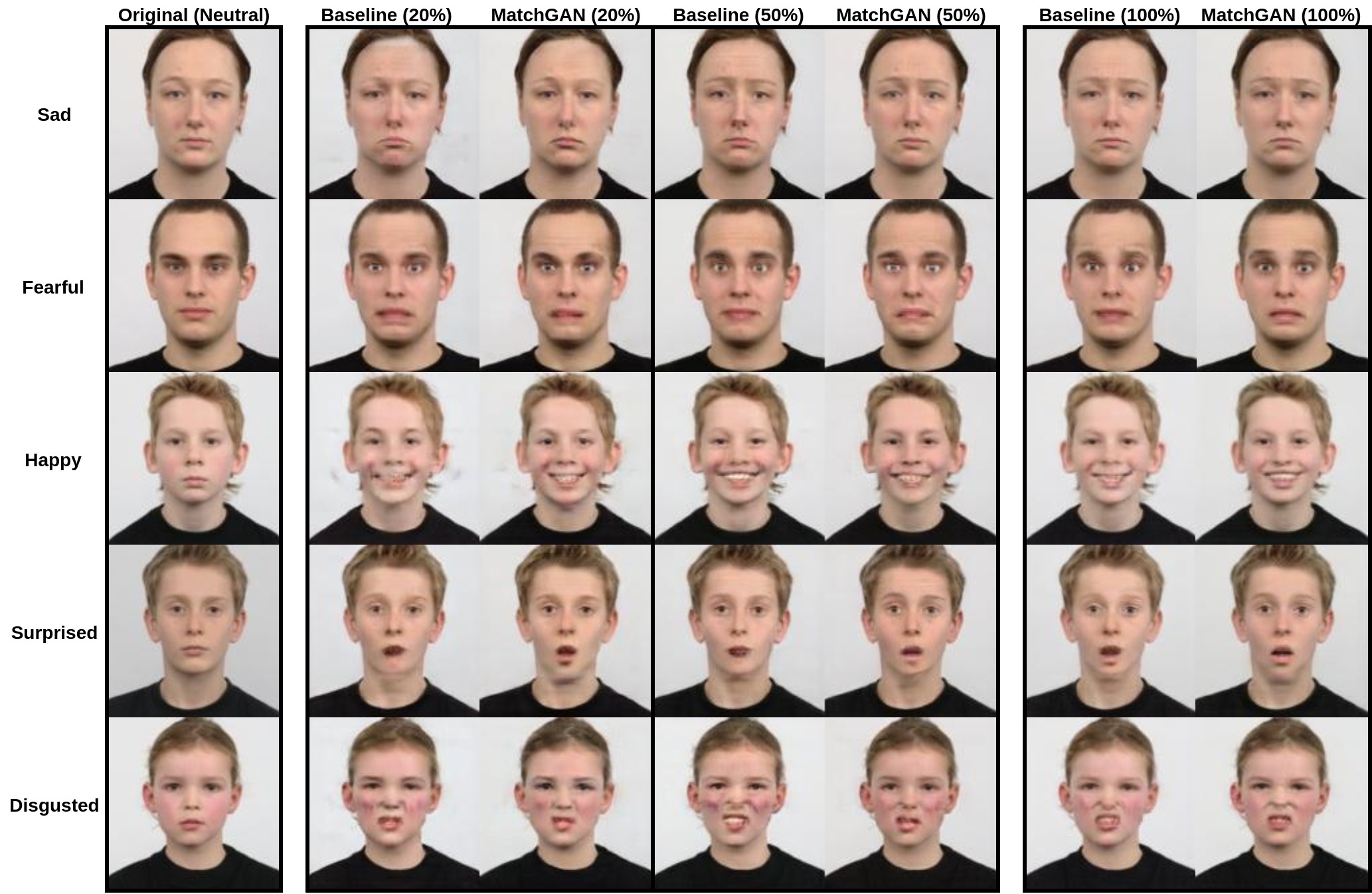}
    \caption{Synthetic examples of MatchGAN vs Baseline on RaFD (zoom in for a better view). Each row corresponds to a single expression manipulation, with target expression listed on the left side.}
    \label{fig:qualitative_rafd}
\end{figure}

Figure \ref{fig:qualitative_celeba} and \ref{fig:qualitative_rafd} compare the visual quality of the images generated by Baseline and MatchGAN on CelebA and RaFD respectively. MatchGAN can be observed to produce images that are less noisy, less blurry, and more coherent. For instance, in the 1\% setup in Figure \ref{fig:qualitative_celeba}, the baseline can often be observed to produce artefacts, blurry patches, or incomplete translations (e.g. the brown patch in the hair in the fourth row) which are not present in the images generated by MatchGAN. Similarly on RaFD, MatchGAN generates more coherent expressions compared to the baseline (e.g. the ``surprised mouth'' in the fourth row in Figure \ref{fig:qualitative_rafd}) and produces fewer artefacts. The image quality of our method also improves substantially with more labelled examples. In Figure \ref{fig:qualitative_celeba}, the overall quality of the images generated by MatchGAN in the 20\% setup is on par with or even outmatches that of the Baseline under the 100\% setup in terms of clarity, colour tone, and coherence of target attributes, corroborating our quantitative results shown in Table \ref{tab:fid_is}.

\section{Conclusion}
In this paper we proposed MatchGAN, a novel self-supervised learning approach for training conditional GANs under a semi-supervised setting with very few labelled examples. MatchGAN utilises synthetic examples and their target labels as additional annotated examples and minimises a triplet matching objective as a pretext task. With 20\% of the training data labelled, it is able to outperform the baseline trained with 100\% of examples labelled and shows a distinct advantage over other self-supervised approaches such as \cite{chen2019self} under both fully-supervised and semi-supervised settings.

\noindent\textbf{Acknowledgements.} This work is supported by the Huawei Consumer Business Group, Croucher Foundation, and EPSRC Programme Grant ‘FACER2VM’ (EP/N007743/1).

\clearpage
\bibliographystyle{splncs}
\bibliography{biblio}

%this would normally be the end of your paper, but you may also have an appendix
%within the given limit of number of pages
\end{document}